\documentclass{article}

\usepackage{PRIMEarxiv}

\usepackage[utf8]{inputenc} 
\usepackage[T1]{fontenc}    
\usepackage{hyperref}       
\usepackage{url}            
\usepackage{booktabs}       
\usepackage{amsfonts}       
\usepackage{nicefrac}       
\usepackage{microtype}      
\usepackage{lipsum}
\usepackage{fancyhdr}       
\usepackage{graphicx}       
\graphicspath{{media/}}     

\usepackage{amsthm}
\usepackage{amsmath}
\usepackage{algorithm}
\usepackage{algpseudocode}
\usepackage{setspace}
\usepackage{makecell}
\usepackage[export]{adjustbox}
\usepackage{xcolor, soul}
\usepackage{subcaption} 
\usepackage{enumitem}
\usepackage{caption}
\usepackage{mathrsfs, amsthm}
\usepackage{amssymb}
\usepackage{bbm,enumerate}
\usepackage{fullpage}

\numberwithin{equation}{section}

\newtheorem{thm}{Theorem}[section]

\newtheorem{claim}[thm]{Claim}

\title{Closing the Gap Between Synthetic and Ground Truth Time Series Distributions via Neural Mapping}

\author{
  Daesoo Lee \\
  Norwegian University of Science and Technology \\
  HANCE
   \And
  Sara Malacarne \\
  Telenor Research \\
  \And
  Erlend Aune \\
  Norwegian University of Science and Technology \\ 
  HANCE \\ 
}

\begin{document}
\maketitle

\begin{abstract}
In this paper, we introduce Neural Mapper for Vector Quantized Time Series Generator (NM-VQTSG), a novel method aimed at addressing fidelity challenges in vector quantized (VQ) time series generation. VQ-based methods, such as TimeVQVAE, have demonstrated success in generating time series but are hindered by two critical bottlenecks: information loss during compression into discrete latent spaces and deviations in the learned prior distribution from the ground truth distribution. These challenges result in synthetic time series with compromised fidelity and distributional accuracy.
To overcome these limitations, NM-VQTSG leverages a U-Net-based neural mapping model to bridge the distributional gap between synthetic and ground truth time series. To be more specific, the model refines synthetic data by addressing artifacts introduced during generation, effectively aligning the distributions of synthetic and real data. Importantly, NM-VQTSG can be used for synthetic time series generated by any VQ-based generative method.
We evaluate NM-VQTSG across diverse datasets from the UCR Time Series Classification archive, demonstrating its capability to consistently enhance fidelity in both unconditional and conditional generation tasks. The improvements are evidenced by significant improvements in FID, IS, and conditional FID, additionally backed up by visual inspection in a data space and a latent space.
Our findings establish NM-VQTSG as a new method to improve the quality of synthetic time series.
Our implementation is available on \url{https://github.com/ML4ITS/TimeVQVAE}.
\end{abstract}


\section{Introduction}

In the field of time series generation (TSG), methodological advances have been significantly shaped by three foundational generative models \cite{ang2023tsgbench}: Generative Adversarial Networks (GAN) \cite{goodfellow2020generative}, Variational AutoEncoders (VAE) \cite{kingma2013auto}, and Vector Quantized Variational AutoEncoders (VQVAE) \cite{van2017neural}. 
GANs, comprising a generator and a discriminator, have been pivotal in TSG for many years \cite{ang2023tsgbench}. They utilize architectures like RNN, LSTM, and Transformer to capture temporal data, with RGAN \cite{esteban2017real} and TimeGAN \cite{yoon2019time} being notable examples. RGAN leverages RNNs within GAN, focusing on the statistical distinction between real and generated series, whereas TimeGAN emphasizes encoding temporal dependencies. 
In contrast to the long-standing dominance of GANs, different approaches have been recently proposed, adopting VAE and VQVAE.
VAEs leverage variational inference to capture a target prior distribution. TimeVAE \cite{desai2021timevae} emerges as a significant development.
Unlike VAEs, which produce a continuous latent space, VQVAEs discretize a latent space, producing a discrete latent space while discarding the Gaussian prior distribution assumption. VQVAEs allow for improving the model's ability to capture a prior distribution and create more crisp data, leading to improved fidelity in the generated outputs. 
TimeVQVAE \cite{lee2023vector}, a specialized adaptation of VQVAE, is specifically designed for TSG. \cite{ang2023tsgbench,lee2023vector} clearly demonstrated that TimeVQVAE and TimeVAE dethroned the GAN-based methods. 
In this paper, we focus on VQ-based TSG methods like TimeVQVAE. It should be noted that the limitations and proposal addressed in this paper are not specific to TimeVQVAE but to any TSG methods based on VQ. At the time of writing, TimeVQVAE is the only VQ-based TSG method, thereby used as a representative for VQ-based TSG methods.

\begin{figure}[!ht]
    \centering
    \includegraphics[width=0.4\textwidth]{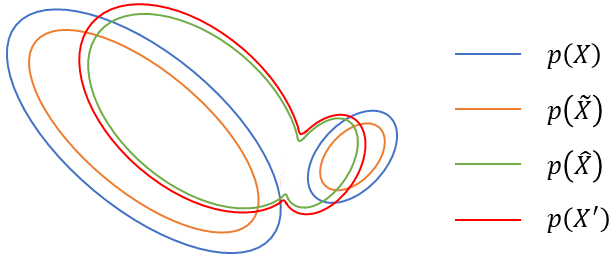}
    \caption{Illustration of the two bottlenecks that cause a discrepancy between the distributions of GT samples $X$ and generated samples $\hat{X}$. 
    $p(X)$, $p(\tilde{X})$, and $p(\hat{X})$ denote a distribution of GT samples $X$ (blue), reconstructed samples $\tilde{X}$ (orange), and generated samples $\hat{X}$ (green), respectively. 
    Lastly, $p(X^\prime)$ represents a distribution of a stochastic variant of $X$ with stochastic vector quantization (red). 
    }
    \label{fig:problem_statement}
\end{figure}

VQ-based TSG methods, however, are not without their flaws. We identified two bottlenecks that compromise the fidelity of the generated time series. The first bottleneck is the compression process from the data space to a discrete latent space, which can result in information loss. The second bottleneck is the learned prior distribution, which often deviates from a true underlying distribution. 
Fig.~\ref{fig:problem_statement} illustrates the distributional discrepancies between the distributions of ground truth (GT) samples and generated samples, caused by the two bottlenecks. 
As for the first bottleneck, $p(\tilde{X})$ captures a subset of $p(X)$ due to the information loss in the compression process. The distributional gap depends on the compression rate -- the higher rate results in the higher gap and vice versa. From the prior learning perspective, a high compression rate is favored, therefore the gap inevitably exists to some extent. 
Regarding the second bottleneck, $p(\hat{X})$ represents a learned prior distribution. Ideally, this should be close to $p(X)$, however, that is often very challenging and limited by several factors. For instance, a prior model is trained on the discrete latent space which is a compressed space thereby hindering a complete capture of a GT distribution. Other factors involve the prior model size, training steps, regularization techniques, dataset size, and so on. Insufficient prior learning can easily lead to inappropriate bias learned within the model, leading to an inaccurate capture of the GT distribution, as shown in the figure with the mismatch between $p(X)$ and $p(\hat{X})$. 

In this paper, we propose training a neural mapping model to map the distribution of the synthetic samples to match the distribution of GT samples. In other words, the mapping model narrows the gap between $p(\hat{X})$ and $p(X)$ by mapping $\hat{X}$ close to $X$ with a learned mapping model.
We observed that the generated samples contain certain patterns unique to synthetic data, which are absent in GT samples. The mapping model is trained to identify and refine these synthetic-specific patterns, transforming a synthetic sample into one that fits within the GT distribution.
For this mapping model, we utilize U-Net \cite{ronneberger2015u} due to its renowned effectiveness in mapping problems \cite{isola2017image}. While other mapping models could be explored, our primary focus is on demonstrating the feasibility of our proposed approach. Therefore, the exploration of alternative mapping models is left for future research.

The proposed mapping model was validated by comparing the fidelity of generated time series from TimeVQVAE (baseline) with and without the mapping model. Our experimental results indicate that our mapping model demonstrates substantial fidelity improvements across diverse datasets from the UCR Time Series Classification archive \cite{dau2019ucr}. To be more specific, the evaluation was conducted with metrics such as Frechet Inception Distance (FID) \cite{heusel2017gans}, Inception Score (IS) \cite{salimans2016improved}, and conditional Frechet Inception Distance (cFID) \cite{benny2021evaluation}, and the mapping model achieved around 151\% average improvement in FID, 7\% in IS, and 37\% in cFID. The effectiveness is further validated by visual inspections in both data and latent spaces, showcasing improved alignment between synthetic and ground truth time series distributions.

Our contributions consist of:
\begin{itemize}[left=1em,label=\tiny$\bullet$,itemsep=0em,topsep=0ex]
    \item \textit{Introduction of NM-VQTSG}: a novel method using U-Net-based neural mapping to refine synthetic time series distributions,
    \item \textit{Addressing Key Challenges}: The method mitigates two critical issues in VQ-based time series generation: compression-related information loss and a mismatch between a ground truth prior distribution and the learned distribution,
    \item \textit{Broad Applicability}: While demonstrated with TimeVQVAE, the approach is generalizable to other VQ-based time series generation methods, paving a new pathway for addressing fidelity challenges in the domain.
\end{itemize}

\section{Background}

\subsection{VQVAE}

It is a VAE that uses VQ to obtain a discrete latent representation \cite{van2017neural}. It differs from classical VAEs by mapping inputs to a discrete latent space and learning the prior, which leads to avoiding posterior collapse and producing sharp and crisp samples. 
It adopts a two-stage modeling approach, involving tokenization (stage~1) and prior learning (stage~2) with learnable models. 

To be more specific, the stage~1 involves training an encoder $E$ and decoder $D$ by minimizing a reconstruction loss $\mathcal{L}_{rec} = \mathbb{E}_x \| x - \tilde{x} \|$ where $\tilde{x} = D(z_q)$, $z_q = Q(z)$, $z = E(x)$ and $Q$ is a vector quantizer.
The quantizer transforms continuous latent vectors into discrete latent vectors using a codebook $\mathcal{Z}$, in which a discrete latent vector is one of the vectors in $\mathcal{Z}$ and is often termed as \textit{token} referring to its codebook index.
$\mathcal{Z}$ consists of $K$ discrete tokens $\mathcal{Z} = \{z_k\}_{k=1}^K$, where $z_k \in \mathbb{R}^d$ and $d$ denotes the dimension size. 
The quantization process is defined as
\begin{equation}
    \label{eq:z_q}
    (z_q)_{n} = \operatorname*{argmin}_{z_k \in \mathcal{Z}} \|z_{n} - z_k\|,
\end{equation}
where $\mathbb{R}^{d \times N} \ni z =(z_n)_n$ denotes the activation map after the encoder, with $N$ representing a spatial size, $n \in N$ and $z_n \in \mathbb{R}^d$. 
To simplify the notation, $z_q$  can be denoted by a token sequence $s = (s_n)_n$, where
\begin{equation}
\label{eq:element_of_s}
s_{n} = k, \:\:\: \textrm{whenever} \:\:\: (z_q)_{n} = z_k.
\end{equation}
This notation allows to 
 use $D(z_q)$ and $D(s)$ interchangeably.

Stage~2 involves training a prior model on $s$ to learn $p(s)$. There are mainly two options for the model: 1) autoregressive models, 2) non-autoregressive (\textit{i.e.,} bidirectional) models. The latter ones have shown significantly better results than the former ones \cite{chang2022maskgit}.
Given the clear superiority of the bidirectional sampling over the autoregressive approach, this paper will focus on bidirectional prior models for our proposed method and experiments.

\subsection{TimeVQVAE}

It adopts the stage~1 from VQVAE \cite{van2017neural} and stage~2 and the sampling process from MaskGIT \cite{chang2022maskgit}. The novel differences are discrete latent space modeling of a time-frequency domain and low-frequency (LF) and high-frequency (HF) separation.
To be more precise, in stage~1, a time series is first projected into a time-frequency domain using Short-Time Fourier Transform and split into LF and HF bands. Then, the processed input is compressed into a set of tokens by an encoder and vector quantizer in both LF and HF bands. Those tokens can be decoded back to a data space with a decoder. 
In stage~2, tokens are randomly masked and a prior model is trained by predicting the GT tokens before the masking. Two bidirectional prior models (\textit{i.e.,} transformer) are employed, one for LF and the other for HF. 
The sampling process involves initially sampling LF tokens, followed by HF tokens conditioned on the sampled LF tokens. Subsequently, the tokens can be decoded and summed to produce a synthetic time series.

\subsection{Limitation of VQ-based TSG Methods}

The sampling process of both autoregressive and non-autoregressive models is iterative. In each iteration, one or multiple tokens are sampled, and this process continues until an entire token sequence is sampled. However, due to the stochastic nature of sampling, there is always a risk of sampling inappropriate tokens. The primary issue is that once a token is sampled, it cannot be retracted, leading to a sampled token sequence $\hat{s}$ that may not fit within the GT distribution $p(s)$, resulting in a synthetic time series with lower fidelity.

We acknowledge that $\hat{s}$ can inevitably contain inappropriate tokens, leading to undesirable patterns in the synthetic time series $\hat{x} = D(\hat{s})$. Consequently, instead of improving the sampling process, our focus is on refining the produced $\hat{x}$ using a neural mapping model to improve its fidelity, making it appear more realistic.

\section{Method}

\begin{thm}
\label{theorem:final}
There exists a learnable mapping 
$p_\theta(\hat{X}) \rightarrow p(X)$, for $\theta$ learnable parameters of a prior model $f_\theta$,  that performs $\hat{X} \rightarrow X$. 
\end{thm}
\begin{proof}
%
The mapping $\hat{X} \rightarrow X$ can be expressed as $p_\theta(\hat{X}) \rightarrow p(X)$ where the mapping function $\rightarrow$ is estimated by a neural mapping model. For clarity, we specify the notation $p(\hat{X})$ as $p_\theta(\hat{X})$, where $\theta$ represents the learnable parameters of a prior model.
Training the mapping model requires pairs of $(\hat{x},x)$, however, such pairs cannot exist since $\hat{x}$ is independently sampled with a generative model. 
To enable the training, this theorem demonstrates that $\hat{X}$ can be approximated by a stochastic variant of the training dataset, denoted by $\tilde{X}^\prime$, so that $\hat{X} \approx \tilde{X}^\prime$ and pairs of $\left( \tilde{x}^\prime,x \right)$ can be used for training the neural mapping model.

Let $\tilde{x}^\prime$ denote a stochastic variant of $x$. Then it can be sampled using the following, 
\begin{subequations}
\begin{equation}
\tilde{x}^\prime  
= D({z}_q^\prime) 
= D(Q^\prime({z},\tau)) 
= D(Q^\prime(E({x}),\tau)),
\end{equation}
\begin{equation}
\label{eq:svq}
z_q^\prime \sim Q^\prime(z,\tau) = \operatorname*{softmax}_{z_k \in \mathcal{Z}}(-\| z_n - z_k \|_2 / \tau) \; \forall n,
\end{equation}
\end{subequations}
where $\tau$ is a temperature parameter to control the entropy of the $\text{softmax}$ distribution, $E$ is an encoder, and $D$ represents a decoder.
Eq.~\eqref{eq:svq} appears similar to Eq.~\eqref{eq:z_q} except that it stochastically samples a token based on the negative Euclidean distances instead of using the argmin process. We refer to Eq.~\eqref{eq:svq} as stochastic vector quantization (SVQ), denoted by $Q^\prime$ which allows the notation, $z_q^\prime \sim Q^\prime(z, \tau)$. 
The variance of the SVQ process is controlled by $\tau$, in which $0 < \tau < 1$ decreases the entropy and $1 < \tau$ increases it, leading to a higher sample diversity.

\begin{claim}
\label{claim}
We claim that $p(\hat{X}) \approx p(\tilde{X}^\prime)$ for an optimal choice of $\tau$.
\end{claim}

Or equivalently, in the discrete latent space, $p_\theta(\hat{S}) \approx p_\phi(S^\prime)$, where $S^\prime$ denotes sequences of the stochastically sampled tokens of $X$ and $\phi$ represents the parameters of $E$ and $D$.
Firstly, $\hat{s}$ consists of $\{ \hat{s}_1, \hat{s}_2, \cdots, \hat{s}_n, \cdots, \hat{s}_N \}$ and $\hat{s}_n$ is sampled by $p_\theta ( \hat{s}_n | \hat{s}_{M_t} )$ for bidirectional prior models, where $\hat{s}_{M_t}$ denotes a (partially) masked token sequence with a masking ratio determined by a sampling iteration step $t$. That is, a new token is sampled given the sampled tokens from the previous iterations. 
Then, the likely tokens for $\hat{s}_n$ tend to capture similar data patterns when the conditional context $\hat{s}_{M_t}$ remains the same, as demonstrated in \cite{lee2024explainable}\footnote{It presents multiple samples of $\hat{x} = D(\hat{s})$ where $\hat{s} \sim p_\theta(\hat{s}_n | \hat{s}_{M_t})$ for any $n$ with the same $\hat{s}_{M_t}$, and the resulting samples show similar data patterns.}. 
Secondly, the distribution in  Eq.~\eqref{eq:svq} is defined based on the negative Euclidean distances where the distance represents a pattern dissimilarity -- larger distance generally equals larger dissimilarity, and vice versa \cite{lee2023masked}. This indicates that $\tilde{X}^\prime$ captures similar data patterns as $X$ and the dissimilarity between $\tilde{X}^\prime$ and $X$ can be increased by increasing $\tau$. 
Based on these two observations, an optimal $\tau$ can be determined such that the dissimilarity level between $X$ and $\tilde{X}^\prime$ matches the dissimilarity between $X$ and $\hat{X}$.

Finally, given Claim~\ref{claim} and the considerations above, the mapping $p_\theta(\hat{X}) \rightarrow p_\phi(X)$ can be learned by 
$$\operatorname*{argmin}_{\varphi}{\mathbb{E}_{x,\tilde{x}^\prime \sim D(Q^\prime(E(x),\tau))} \| x - f_{\varphi}(\tilde{x}^\prime) \|},$$
where $\varphi$ represent the learnable parameters of the mapping model $f_\varphi$.
\end{proof}

One aspect we do not account for is the varying variances in $p_\theta(\hat{s}_n | \cdot)$ across the different sampling iteration steps. Typically, the variance is larger at the beginning of the sampling process and decreases in the later steps. Regardless, we use a single $\tau$ in Eq.~\eqref{eq:svq} instead of using $\tau = \{ \tau_1, \tau_2, \cdots, \tau_n, \cdots, \tau_N \}$ to simplify our proposal. We leave this to future work.

An optimal $\tau$ is obtained by calculating FID between $\hat{X}$ and $X^\prime$ with ROCKET \cite{dempster2020rocket} as a feature extractor for a range of different values of $\tau$, and select $\tau$ that results in the lowest FID as the optimal $\tau$. ROCKET is adopted because its representations are universal and robust, and most importantly it does not require any training, allowing its use out of the box. To add robustness in FID, we normalize the ROCKET's representations with L2 norm. 
The search range of \{0.1, 0.5, 1, 2, 4\} for $\tau$ is used in our experiments. A more sophisticated search algorithm could be adopted, however we leave that to future work since we found this simple approach sufficient.

Lastly, $f_\varphi$ is trained by minimizing the following loss function: ${\mathbb{E}_{x,\tilde{x}^\prime \sim D(Q^\prime(E(x),\tau))} \| x - f_\varphi(\tilde{x}^\prime) \|_1}$ (\textit{i.e.,} $L_1$ loss) in this study. Pseudocode for the training process is presented in Algorithm~\ref{alg:training}. While more sophisticated loss functions could be employed, we defer this to future work and suggest a potential candidate in the discussion section.
Then, the inference can be performed as $\hat{x}_R = f_\varphi(\hat{x})$, improving the fidelity of a synthetic time series $\hat{x}$ generated by a VQ-based TSG method through the proposed mapping. $\hat{x}_R$ should retain the same context as $\hat{x}$ while improving its fidelity by refining patterns that are specific to the synthetic time series.

We call our proposed method \textit{Neural Mapper for Vector Quantized Time Series Generator} (NM-VQTSG). For brevity, we will use the abbreviation NM in this paper.


\begin{algorithm}[!ht]
\caption{Training process of $f_\varphi$}
\label{alg:training}
\small
\begin{spacing}{1.1}
\begin{algorithmic}

\State $E$, $D$, and $f_\theta$ have been already trained.  \Comment{$f_\theta$ indicates a prior model.}

\\

\State \# Finding an optimal $\tau$
\State Initialize $\tau \leftarrow 0.1$, $m^* \leftarrow \infty$
\State $\hat{X} \leftarrow$ sampling with the trained generative model 
\For{each $\tau_i \in \{0.1, 0.5, 1, 2, 4\}$}  \Comment{This range is used for the candidates of $\tau$ in our experiments.}
    \State $\tilde{X}^\prime \leftarrow D(Q^\prime( E(X), \tau ))$
    \State $m \leftarrow \text{calculate FID between } \hat{X} \text{ and } \tilde{X}^\prime$ \text{with ROCKET}
    \If{$m < m^*$}
        \State $m^* \leftarrow m$
        \State $\tau \leftarrow \tau_i$
    \EndIf
\EndFor

\\

\State \# Training
\State Initialize $f_\varphi$
\For {$\text{training step} \in \{ 1,2,3, \cdots \}$}
    \For {$x \in X$}  \Comment{$X$ is a training set. In practice, mini-batch is used.}
        \State $x^\prime \leftarrow D(Q^\prime(E(x), \tau))$
        \State update $\varphi$ by minimizing $\| x - f_\varphi(x^\prime) \|_1$
    \EndFor
\EndFor

\end{algorithmic}
\end{spacing}
\end{algorithm}





\section{Experiments}

\subsection{Experimental Setup}

\paragraph{Dataset}
We evaluate our approach using datasets from the UCR Time Series Classification Archive, which comprises over 128 distinct time series datasets. To ensure robust and meaningful assessments, we selected the thirteen largest datasets, excluding those that were redundant or exhibited minimal pattern variation. The chosen datasets include Crop, ECG5000, ElectricDevices, FaceAll, FordA, FreezerRegularTrain, MixedShapesRegularTrain, StarLightCurves, TwoPatterns, UWaveGestureLibraryAll, UWaveGestureLibraryX, Wafer, and Yoga. Each dataset was reconfigured by merging and subsequently performing a stratified split, allocating 80\% of the data to the training set and the remaining 20\% to the test set. This ensures sufficiently large training dataset for generative modeling while maintaining a consistent distribution between the training and test datasets. The used datasets in our experiments can be found here\footnote{\url{https://figshare.com/articles/dataset/UCR_Archive_2018_resplit_ver_/26206355?file=47494442}}.

\paragraph{Models}
For TimeVQVAE, we adopted the implementation available here\footnote{\url{https://github.com/ML4ITS/TimeVQVAE}}. The encoder and decoder of TimeVQVAE consist of multiple downsampling and upsampling convolutional layers with several convolutional residual blocks at every depth level. Specifically, each downsampling layer reduces the temporal dimension by half while doubling the feature dimension. The quantizer module projects the feature dimension of the encoder's embeddings down to eight and utilizes a codebook size of 1024, following \cite{yu2022vectorquantized}. Additionally, the prior model incorporates multiple transformer layers with self-attentions, followed by a small prediction head.
As for NM, it has a U-Net architecture with skip connections with its input as time series. Its encoder and decoder architectural designs are similar to those of TimeVQVAE; however, the NM’s U-Net uses 1-dimensional convolutional layers and integrates self-attention layers alongside the convolutional residual blocks. The full details are provided in the appendix.

\paragraph{Training Procedure}
There are three training stages: stage~1 (tokenization), stage~2 (prior learning), and stage~3 (training NM).
For all the stages, we employed an initial learning rate of 0.005, which decayed to 0.0005 using a cosine learning rate schedule with a linear warmup. The effective batch size was set to 32. 
For the number of training steps, different values are used for different stages -- 20,000 for stage~1, 40,000 for stage~2, and 30,000 for stage~3.
All models were implemented in PyTorch and optimized using the AdamW optimizer \cite{loshchilov2018decoupled}.

\paragraph{Metrics}
We evaluate the models’ generative performance for unconditional generation with FID and IS, as suggested in \cite{lee2023vector}, using a pretrained supervised fully convolutional network (FCN) from here\footnote{\url{https://github.com/danelee2601/supervised-FCN-2}} -- this model was trained on the same training dataset employed in this study. 
For class-conditional generation, we use cFID, referred to as WCFID in \cite{benny2021evaluation}. cFID is computed by averaging the FID scores for each class. Classification Accuracy Score (CAS) \cite{chang2022maskgit,lee2023vector} is sometimes used for this evaluation, but we found this metric difficult to be consistent since it varies much depending on optimizer settings due to the small dataset size of the UCR archive compared to common image benchmark datasets.  
The metrics are reported by averaging over three runs.
In addition, we provide comparative visual plots between $X$ and $\hat{X}_R$ in a data space and a latent space. These visual inspections are critical as the existing numeric metrics cannot fully capture the realism of generated time series.

\paragraph{Competing Models}
We compare TimeVQVAE as a baseline and TimeVQVAE with NM (TimeVQVAE+NM) to independently evaluate the effectiveness of NM.

\section{Results}

\begin{figure}[!ht]
    \includegraphics[width=\textwidth, left]{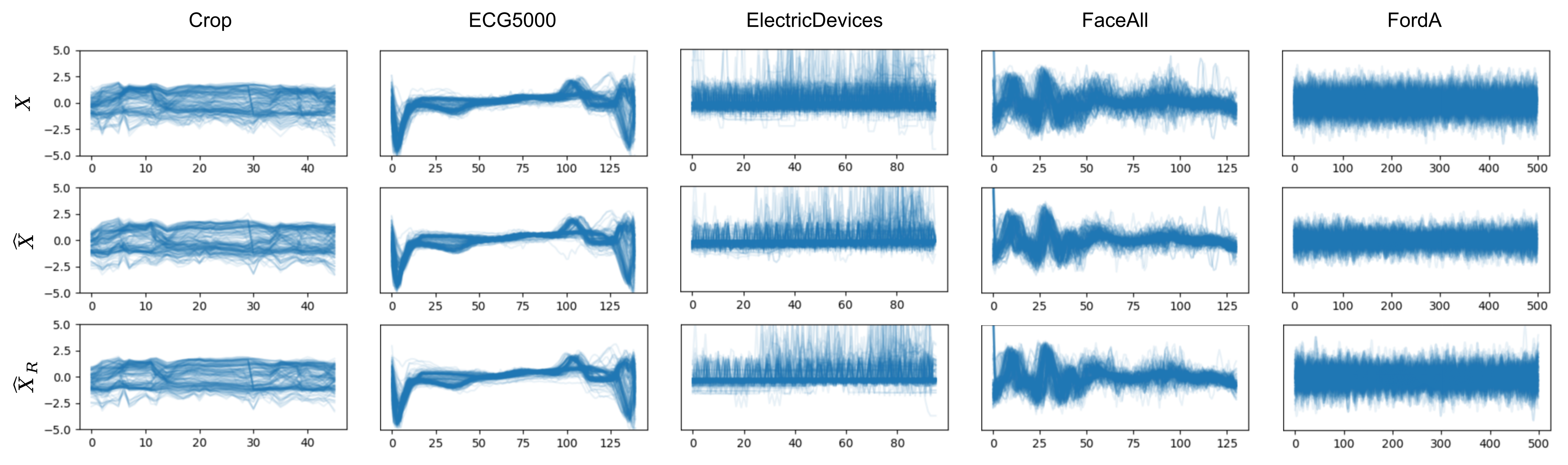} \\[0.5em]
    \includegraphics[width=\textwidth, left]{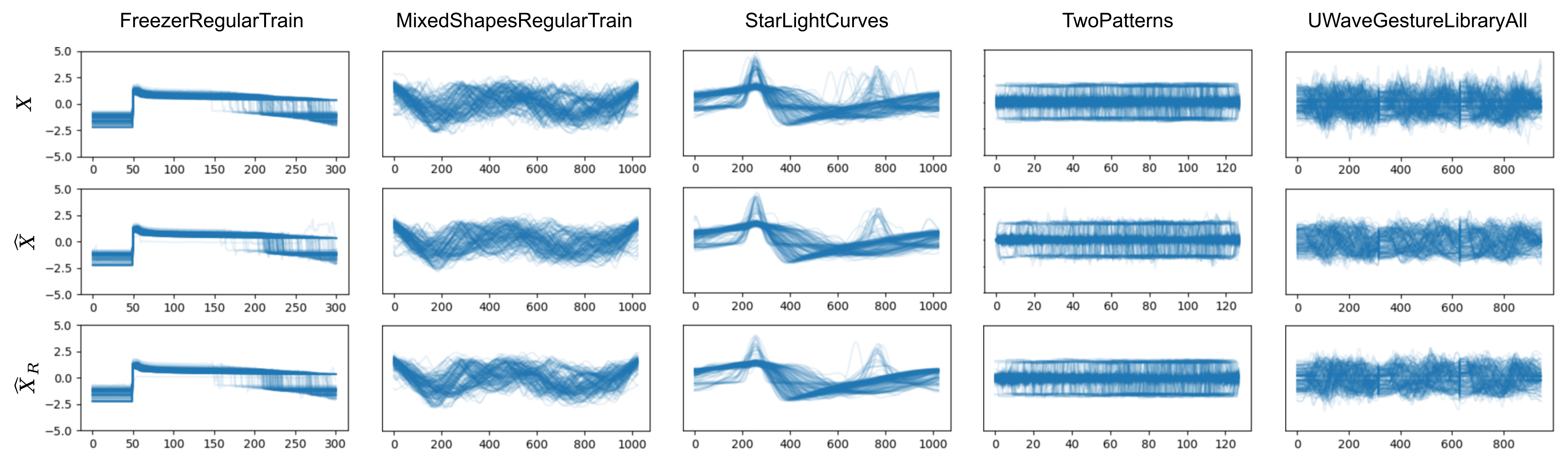} \\[0.5em]
    \includegraphics[width=0.6\textwidth, center]{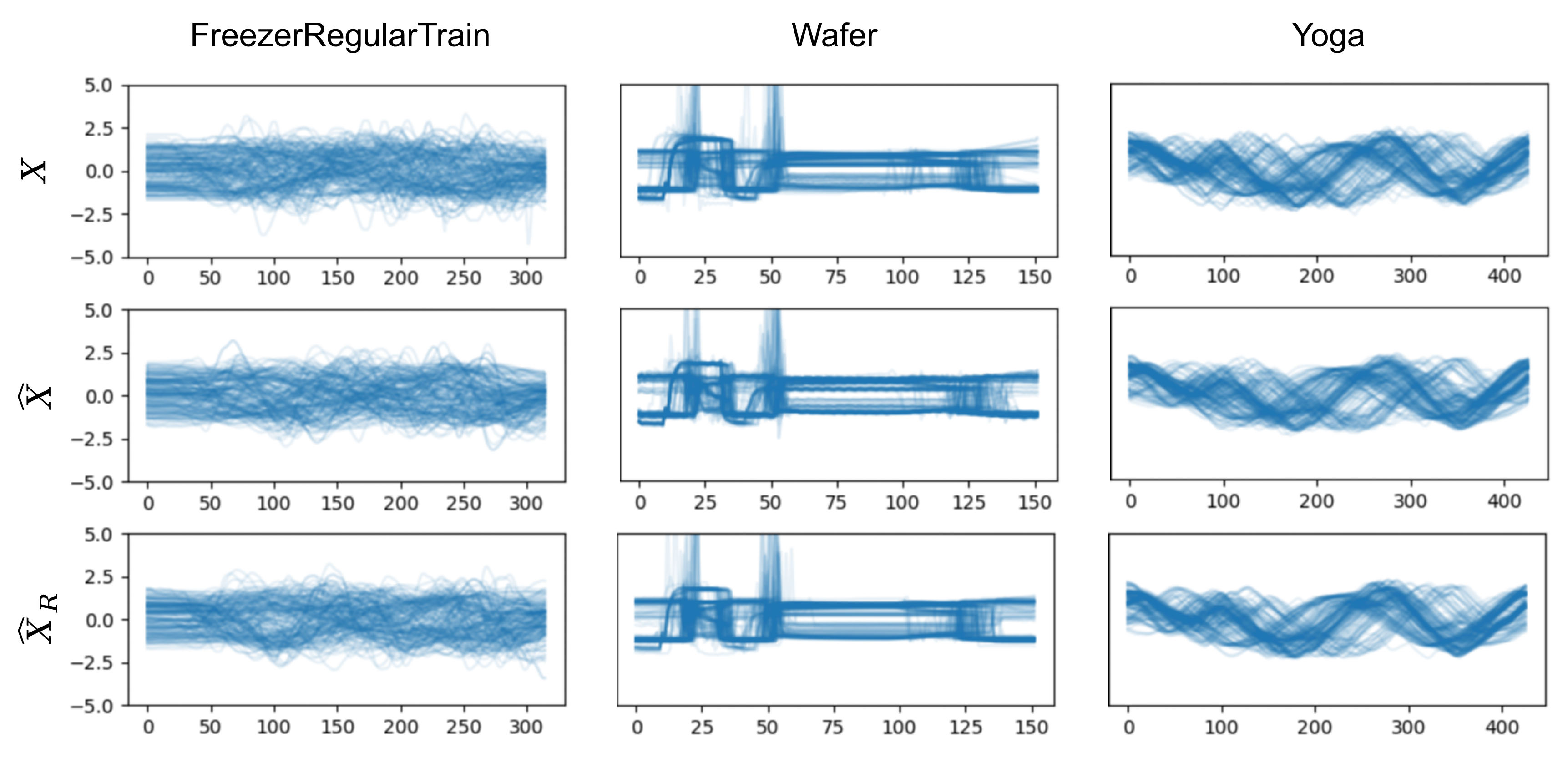}
    \caption{Comparisons between $X$, $\hat{X}$, and $\hat{X}_R$. 200 randomly-sampled time series are plotted in overlap. 
    For FordA and FaceAll, the differences between $X$, $\hat{X}$, and $\hat{X}_R$ are easily observable.
    For FreezerRegularTrain, TwoPatterns, and Wafer, the differences can be better detected in zoom-in, where $\hat{X}$ has somewhat noisy lines and those are corrected in $\hat{X}_R$. 
    }
    \label{fig:X_Xhat_Xhat_R}
\end{figure}

\begin{figure}[!ht]
    \centering
    \includegraphics[width=\textwidth]{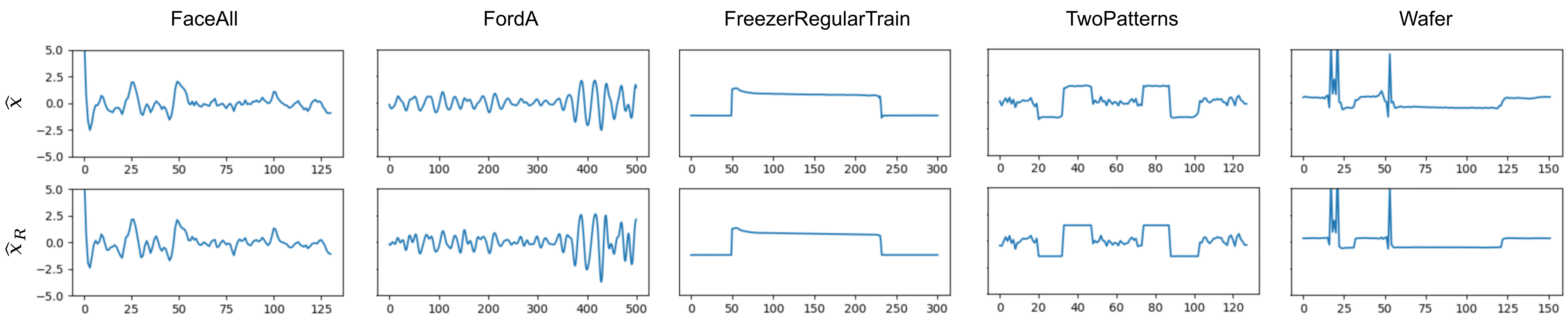}
    \caption{Comparisons between $\hat{x}$ and $\hat{x}_R$. 
    For FaceAll and FordA, the magnitudes are corrected with addition of detailed patterns. 
    For Freezer, a small downward bump at a timestep of around 240 in $\hat{x}$ is corrected in $\hat{x}_R$ (better viewed in zoom-in). 
    For TwoPatterns and Wafer, $\hat{x}$'s horizontal lines are a bit wiggly and those lines are mapped to straight horizontal lines in $\hat{x}_R$ while retaining the rest of the patterns.}
    \label{fig:x_xhat_xhat_R}
\end{figure}

\begin{figure}[!ht]
    \includegraphics[width=\textwidth, right]{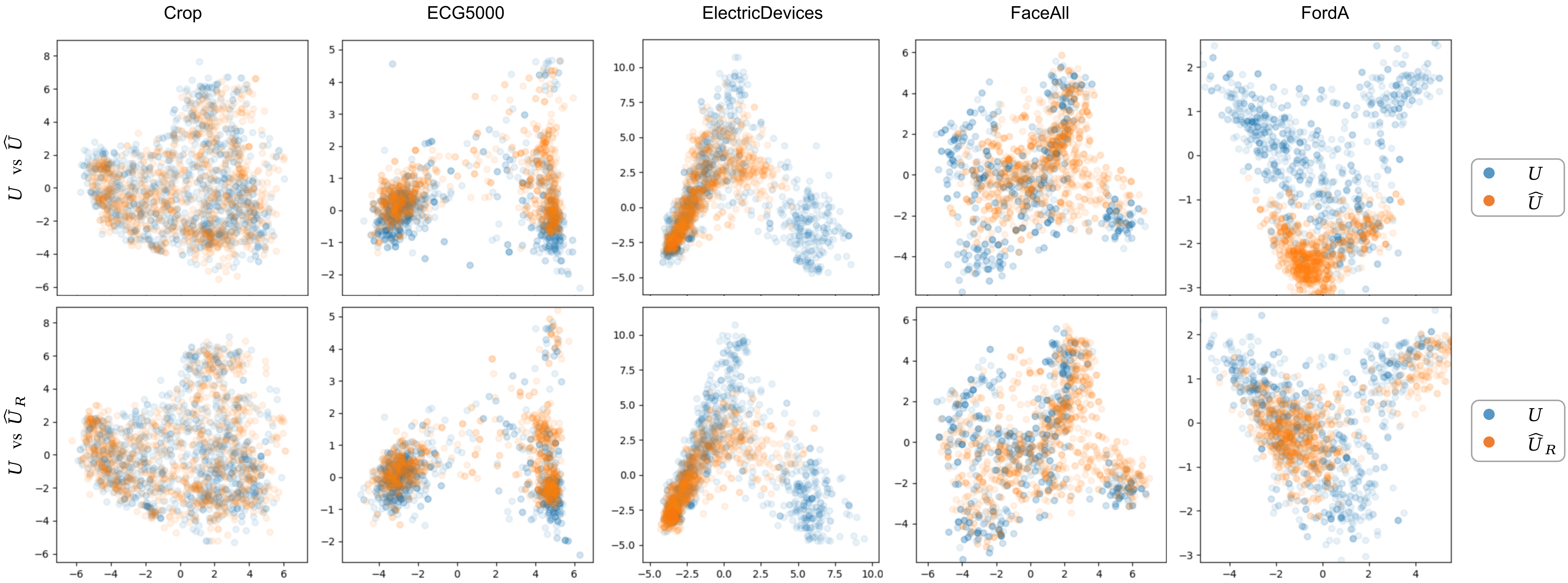} \\[0.5em]
    \includegraphics[width=\textwidth, right]{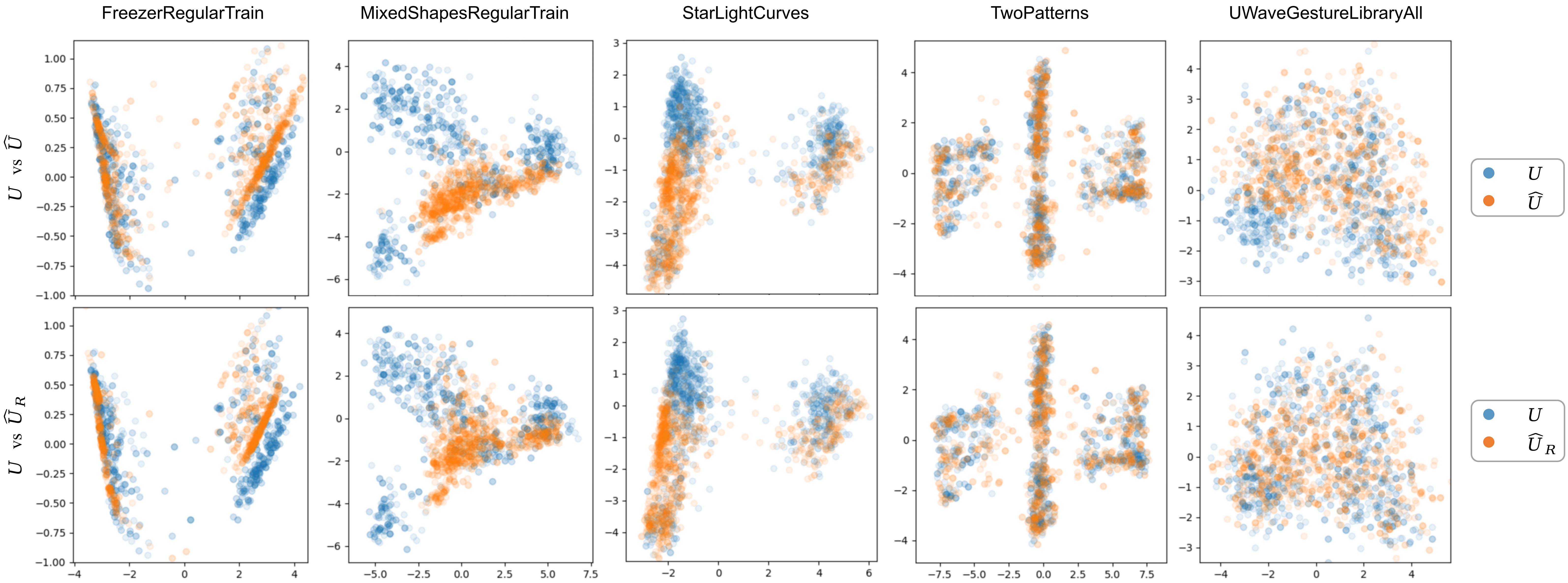} \\[0.5em]
    \includegraphics[width=0.62\textwidth, center]{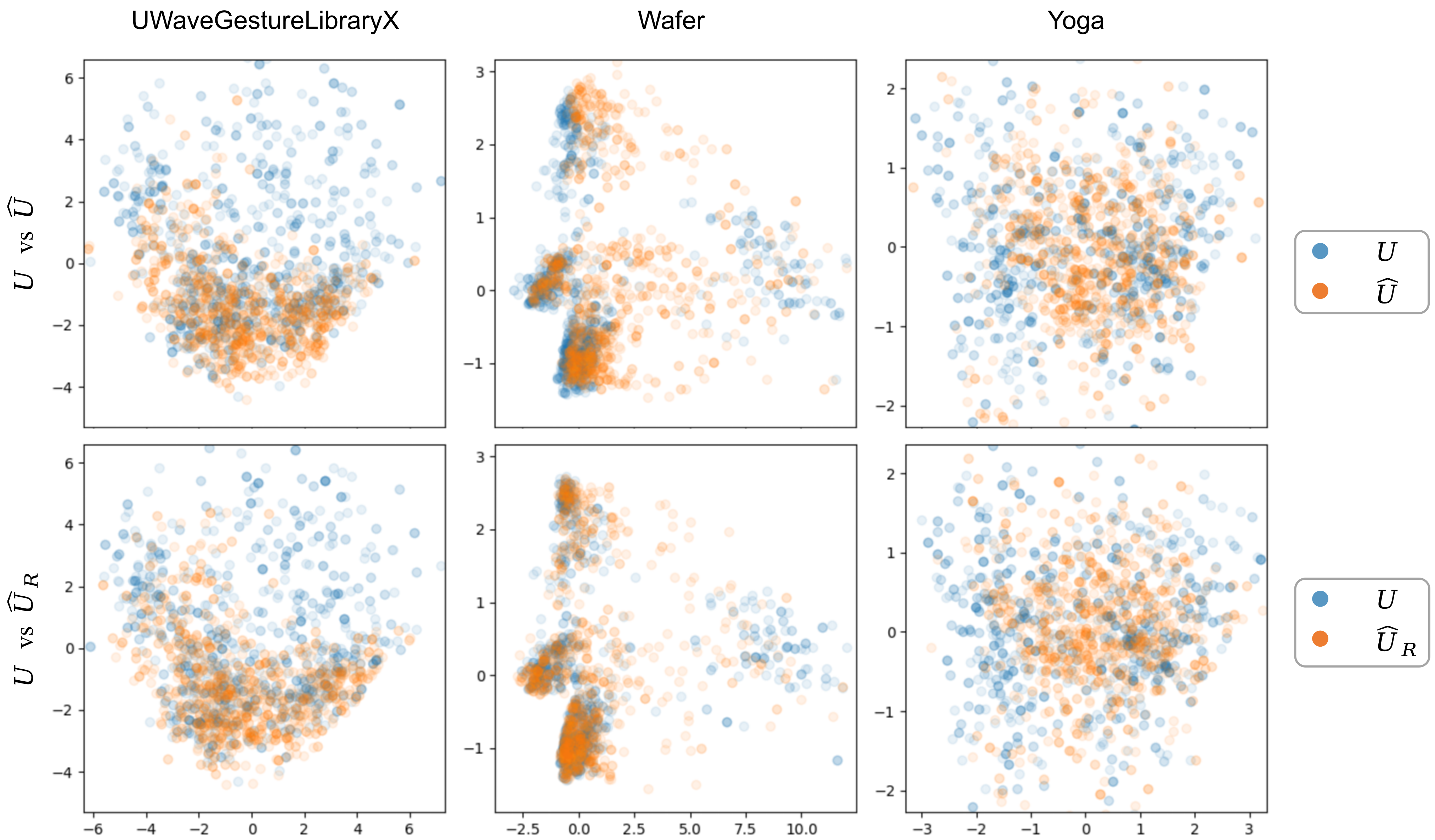}
    \caption{Paired comparisons between $U$ and $\hat{U}$, as well as between $U$ and $\hat{U}_R$. 
    $U$ denotes the representation from the FCN, computed as $U = \text{FCN}(X)$, and similarly for $\hat{U}$ and $\hat{U}_R$ by $\text{FCN}(\hat{X})$ and $\text{FCN}(\hat{X}_R)$, respectively.
    The representations originally have dimension size of 128, but we reduce it to 2 using PCA for the visualization.
    }
    \label{fig:U_Uhat_Uhat_R}
\end{figure}

To evaluate unconditional generation, Tables~\ref{tab:fid} and~\ref{tab:is} present FID and IS, respectively. For conditional generation, Table~\ref{tab:cfid} reports cFID. To visually inspect the generated time series, Fig.~\ref{fig:X_Xhat_Xhat_R} shows comparative plots of $X$, $\hat{X}$, and $\hat{X}_R$, while Fig.~\ref{fig:x_xhat_xhat_R} displays $\hat{x}$ alongside $\hat{x}_R$ for a subset of datasets where realistic patterns are clearly observable. Lastly, Fig.~\ref{fig:U_Uhat_Uhat_R} shows the comparison between $X$, $\hat{X}$, and $\hat{X}_R$ in a latent space.

Overall, the results indicate that NM generally achieves significant fidelity improvements in both unconditional and conditional generation. We provide a detailed analysis of these findings below.

With respect to FID, NM achieves remarkable improvements on datasets such as FordA, showing an increase of 1116.8\%. To investigate this further, we examine Fig.~\ref{fig:X_Xhat_Xhat_R} and Fig.~\ref{fig:x_xhat_xhat_R}. In the FordA dataset, $\hat{X}$ exhibits underestimated magnitudes, whereas $\hat{X}_R$ has magnitudes that are closer to $X$, resulting in the significant improvement in FID. Additionally, Fig.~\ref{fig:U_Uhat_Uhat_R} demonstrates that $X$ and $\hat{X}_R$ are more similar compared to $X$ and $\hat{X}$ for FordA.
Fig.~\ref{fig:U_Uhat_Uhat_R} generally shows that a smaller FID corresponds to a smaller gap between the two distributions in the latent space. However, there are a few datasets where the FID worsens with NM, such as Crop and TwoPatterns. In the case of Crop, we found that $\hat{X}$ is already very close to $X$, leaving little room for fidelity improvement with NM. Consequently, NM cannot further enhance fidelity for Crop. This observation is consistent across other datasets like TwoPatterns and Yoga, where the distributional gap between $X$ and $\hat{X}$ remains minimal.
Although the deterioration of -45\% in Crop may seem substantial, we did not observe a notable discrepancy between $\hat{X}$ and $\hat{X}_R$, highlighting the importance of visual inspection. Specifically, for TwoPatterns, visual inspection using Fig.~\ref{fig:X_Xhat_Xhat_R} and Fig.~\ref{fig:x_xhat_xhat_R} reveals that $\hat{X}_R$ exhibits higher fidelity than $\hat{X}$ by displaying sharp horizontal lines.

Regarding IS, NM delivers considerable improvements across most datasets. However, a slight decline of 11.7\% is observed in the Wafer dataset. We identified that this decrease is due to a class imbalance in the dataset, where 10\% of the samples belong to class~1 and 90\% to class~2. This imbalance results in poor mapping performance for class~1, leading to a degraded IS. As with any model training, addressing class imbalance typically requires appropriate techniques. In this study, we did not implement such techniques to keep the experimental settings simple.

Regarding cFID, similar to FID, substantial improvements are observed overall. However, setbacks occur for the Crop and Wafer datasets due to the limited room for improvement and class imbalance in the dataset, respectively, as previously discussed. Additionally, a setback is noted for FreezerRegularTrain. Despite this, visual inspection did not reveal significant fidelity degradation; in fact, $\hat{X}_R$ is preferred for its sharp lines and appears closer to $X$. We suspect that the cFID scale is so small that even minor differences result in the large change, even though the actual differences are negligible.

\begin{figure}[!ht]
    \centering
    \begin{subfigure}[b]{\textwidth} 
        \centering
        \includegraphics[width=\textwidth]{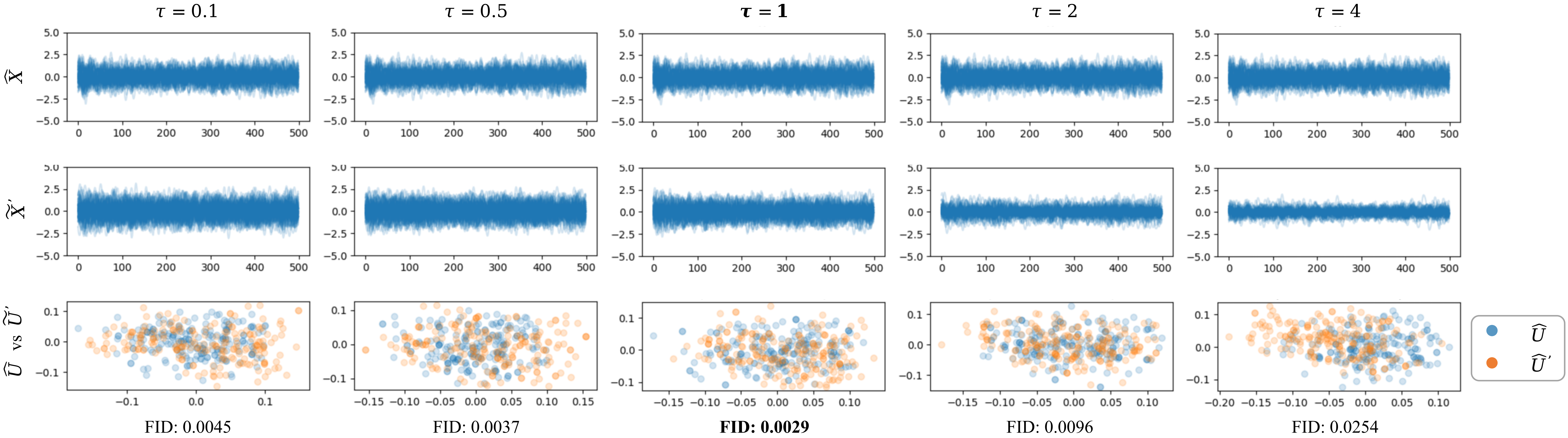}
        \caption{Dataset: FordA}
        \label{fig:sub1}
    \end{subfigure}
    \\[1em]
    \begin{subfigure}[b]{\textwidth} 
        \centering
        \includegraphics[width=\textwidth]{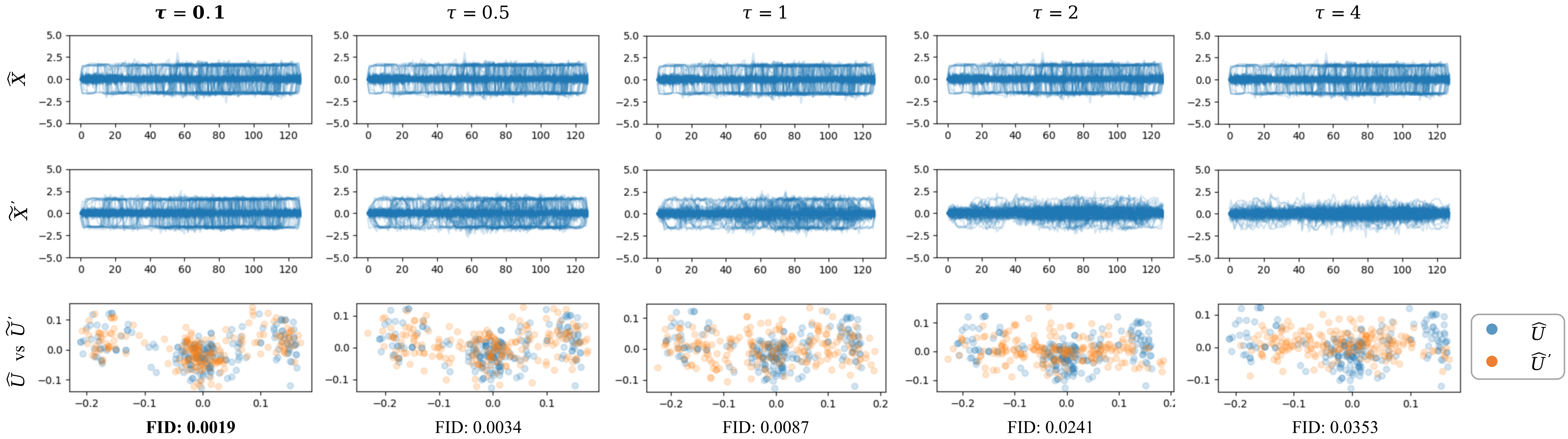}
        \caption{Dataset: TwoPatterns}
        \label{fig:sub2}
    \end{subfigure}
    \caption{Experimental validation that our hypothesis of $p(\hat{X}) \approx p(\tilde{X}^\prime)$ is satisfied with an optimal $\tau$. The optimal $\tau$ is 1 and 0.1 for (a) and (b), respectively.
    $\hat{U}$ denotes the representation from ROCKET, computed as $\hat{U} = \text{ROCKET}(\hat{X})$, and similarly for $\tilde{U}^\prime$ by $\text{ROCKET}(\tilde{X}^\prime)$.
    The original dimension size of the representations is reduced to 2 using PCA for the visualization.
    }
    \label{fig:hypothesis_test}
\end{figure}

Finally, we experimentally validate our hypothesis that \( p(\hat{X}) \approx p(\tilde{X}^\prime) \) by demonstrating that \( \hat{X} \) closely resembles \( \tilde{X}^\prime \) in a latent space using an optimal \( \tau \). To determine the optimal \( \tau \), we select the value that minimizes the FID computed with ROCKET, thereby ensuring that \( \hat{X} \approx \tilde{X}^\prime \) within ROCKET's latent space. Fig.~\ref{fig:hypothesis_test} compares \( \hat{X} \) and \( \hat{X}^\prime \) across different values of \( \tau \) in both data and latent spaces for a subset of the datasets. The results demonstrate two key points: first, the simple search algorithm is sufficient for finding an optimal \( \tau \), and second, our hypothesis is effectively satisfied and validated.

\begin{table}[!ht]
\centering
\small
\caption{FID (the lower, the better performance). "Change" is calculated as: $((\text{FID of TimeVQVAE} / \text{FID of TimeVQVAE+NM}) \times 100) - 100$.}
\label{tab:fid}
\begin{tabular}{lccc}
\toprule
{Dataset Name} & {TimeVQVAE} & \makecell{TimeVQVAE\\+NM} & \makecell{Change\\(\%)} \\
\midrule
Crop & 1.74 & 3.21 & -45.9 \\
ECG5000 & 1.03 & 0.86 & 20.0 \\
ElectricDevices & 19.67 & 18.03 & 9.1 \\
FaceAll & 10.19 & 5.88 & 73.5 \\
FordA & 7.85 & 0.65 & 1116.8 \\
FreezerRegularTrain & 0.29 & 0.10 & 200.5 \\
MixedShapesRegularTrain & 17.98 & 10.22 & 75.8 \\
StarLightCurves & 5.63 & 5.63 & 0.0 \\
TwoPatterns & 0.32 & 0.38 & -17.5 \\
UWaveGestureLibraryAll & 7.38 & 3.65 & 102.3 \\
UWaveGestureLibraryX & 5.40 & 3.47 & 55.4 \\
Wafer & 1.37 & 0.36 & 280.8 \\
Yoga & 0.31 & 0.30 & 5.7 \\
\bottomrule
\end{tabular}
\end{table}

\begin{table}[!ht]
\centering
\small
\caption{IS (the higher, the better performance). \#class denotes a number of classes. "(normalized)" denotes a normalized IS, calculated as IS divided by \#class. "Change" is calculated as: $((\text{IS of TimeVQVAE+NM} / \text{IS of TimeVQVAE}) \times 100) - 100$.}
\label{tab:is}
\begin{tabular}{lcccccc}
\toprule
{Dataset Name} & {\#class} & {TimeVQVAE} & \makecell{TimeVQVAE\\+NM} & \makecell{TimeVQVAE\\(normalized)} & \makecell{TimeVQVAE\\+NM\\(normalized)} & \makecell{Change\\(\%)} \\
\midrule
Crop & 24 & 12.55 & 12.89 & 0.52 & 0.54 & 2.7 \\
ECG5000 & 5 & 2.09 & 2.11 & 0.42 & 0.42 & 0.7 \\
ElectricDevices & 7 & 3.22 & 3.82 & 0.46 & 0.55 & 18.8 \\
FaceAll & 14 & 5.31 & 7.13 & 0.38 & 0.51 & 34.2 \\
FordA & 2 & 1.44 & 1.62 & 0.72 & 0.81 & 12.7 \\
FreezerRegularTrain & 2 & 1.93 & 1.96 & 0.97 & 0.98 & 1.2 \\
MixedShapesRegularTrain & 5 & 2.66 & 3.09 & 0.53 & 0.62 & 16.5 \\
StarLightCurves & 3 & 2.12 & 2.20 & 0.71 & 0.73 & 4.1 \\
TwoPatterns & 4 & 3.76 & 3.77 & 0.94 & 0.94 & 0.3 \\
UWaveGestureLibraryAll & 8 & 4.21 & 4.41 & 0.53 & 0.55 & 4.9 \\
UWaveGestureLibraryX & 8 & 4.33 & 4.55 & 0.54 & 0.57 & 5.2 \\
Wafer & 2 & 1.50 & 1.32 & 0.75 & 0.66 & -11.7 \\
Yoga & 2 & 1.52 & 1.52 & 0.76 & 0.76 & 0.2 \\
\bottomrule
\end{tabular}
\end{table}

\begin{table}[!ht]
\centering
\small
\caption{cFID (the lower, the better performance). "Change" is calculated as: $((\text{cFID of TimeVQVAE} / \text{cFID of TimeVQVAE+NM}) \times 100) - 100$.}
\label{tab:cfid}
\begin{tabular}{lccc}
\toprule
{Dataset Name} & {TimeVQVAE} & \makecell{TimeVQVAE\\+NM} & \makecell{Change\\(\%)} \\
\midrule
Crop & 6.03 & 7.74 & -22.1 \\
ECG5000 & 4.61 & 4.18 & 10.3 \\
ElectricDevices & 25.06 & 24.00 & 4.4 \\
FaceAll & 9.36 & 5.45 & 71.9 \\
FordA & 8.24 & 2.29 & 260.2 \\
FreezerRegularTrain & 0.12 & 0.16 & -25.0 \\
MixedShapesRegularTrain & 21.15 & 12.58 & 68.1 \\
StarLightCurves & 2.82 & 2.31 & 22.3 \\
TwoPatterns & 0.21 & 0.21 & -0.8 \\
UWaveGestureLibraryAll & 9.19 & 4.94 & 86.0 \\
UWaveGestureLibraryX & 7.82 & 5.76 & 35.7 \\
Wafer & 0.55 & 0.83 & -34.0 \\
Yoga & 0.50 & 0.43 & 15.8 \\
\bottomrule
\end{tabular}
\end{table}


\section{Discussion}

\paragraph{Advanced Reconstruction Loss}
In this study, NM is trained with a $L_1$ reconstruction loss. A more advanced loss function can be adopted for further improvement, such as a multi-scale STFT reconstruction loss where STFT denotes Short-Time Fourier Transform. It can be defined as $\sum_i \| \mathcal{S}_i(x) - \mathcal{S}_i({f_\varphi(\tilde{x}^\prime)}) \|$ in which $\mathcal{S}$ represents a complex spectrogram transformation function and the subscript $i$ denotes different parameters of the transformation function such as window length and stride size.

\paragraph{Deficiency of Stochasticity} 
NM performs deterministic one-to-one mapping. However, this might not be suitable for all datasets. For some, $\hat{x}$ could correspond to multiple plausible $x_R$ outcomes. Therefore, adoption of stochasticity in the mapping should be a direction towards the universal mapping model.

\paragraph{GAN Loss}
To train NM, a $L_1$ loss is used. But, this loss is known to have tendency of producing blurred data due to the averaging effect. On the other hand, a GAN loss (also known as a discriminative loss) can help it produce sharp and crisp data. 
The averaging effect arises because point-wise losses encourage the model to find a mean prediction when multiple plausible outputs exist, leading to smooth and indistinct features. In contrast, a GAN loss employs a discriminator network that distinguishes between real and generated samples, compelling the generator (\textit{i.e.,} the mapping model in our study) to produce outputs that are not only accurate in a point-wise sense but also rich in fine-grained details. 
In summary, integrating a GAN loss should enable the mapping model to generate sharper and more realistic outputs, preserving fine-grained details that would otherwise be potentially lost with a $L_1$ loss alone.

\section{Conclusion}
This study introduced the Neural Mapper for Vector Quantized Time Series Generator (NM-VQTSG) to address the fidelity challenges inherent in VQ-based time series generation. By leveraging a U-Net-based mapping model, the proposed method effectively narrows the distributional gap between synthetic and ground truth time series. Through rigorous experimentation on diverse datasets from the UCR archive, NM-VQTSG demonstrated consistent improvements in both unconditional and conditional generation tasks, as evidenced by metrics such as FID, IS, and cFID. Additionally, visual inspections confirmed enhanced realism in generated samples, particularly in datasets where fidelity improvements were most needed.

Overall, NM-VQTSG establishes a new line of study by improving fidelity through mapping, offering a robust method for producing realistic synthetic time series. Future work will explore inclusion of stochastic mapping to handle scenarios with multiple plausible outcomes, and a more sophisticated reconstruction loss and a GAN loss to better capture fine-grained details and further improve fidelity.

\section*{Acknowledgments}
We would like to thank the Norwegian Research Council for funding the Machine Learning for Irregular Time Series (ML4ITS) project (312062). This funding directly supported this research.
We also would like to thank the people who have contributed to the UCR Time Series Classification archive \cite{dau2019ucr} for their valuable contribution to the community.

\section*{Ethical Statement}
No conflicts of interest were present during the research process.

\bibliographystyle{unsrt}  
\bibliography{references}

\newpage
\appendix

\section{Implementation Details}


\subsection{NM-VQTSG}

\begin{figure}[!ht]
    \centering
    \begin{subfigure}[b]{0.45\textwidth} 
        \centering
        \includegraphics[width=\textwidth]{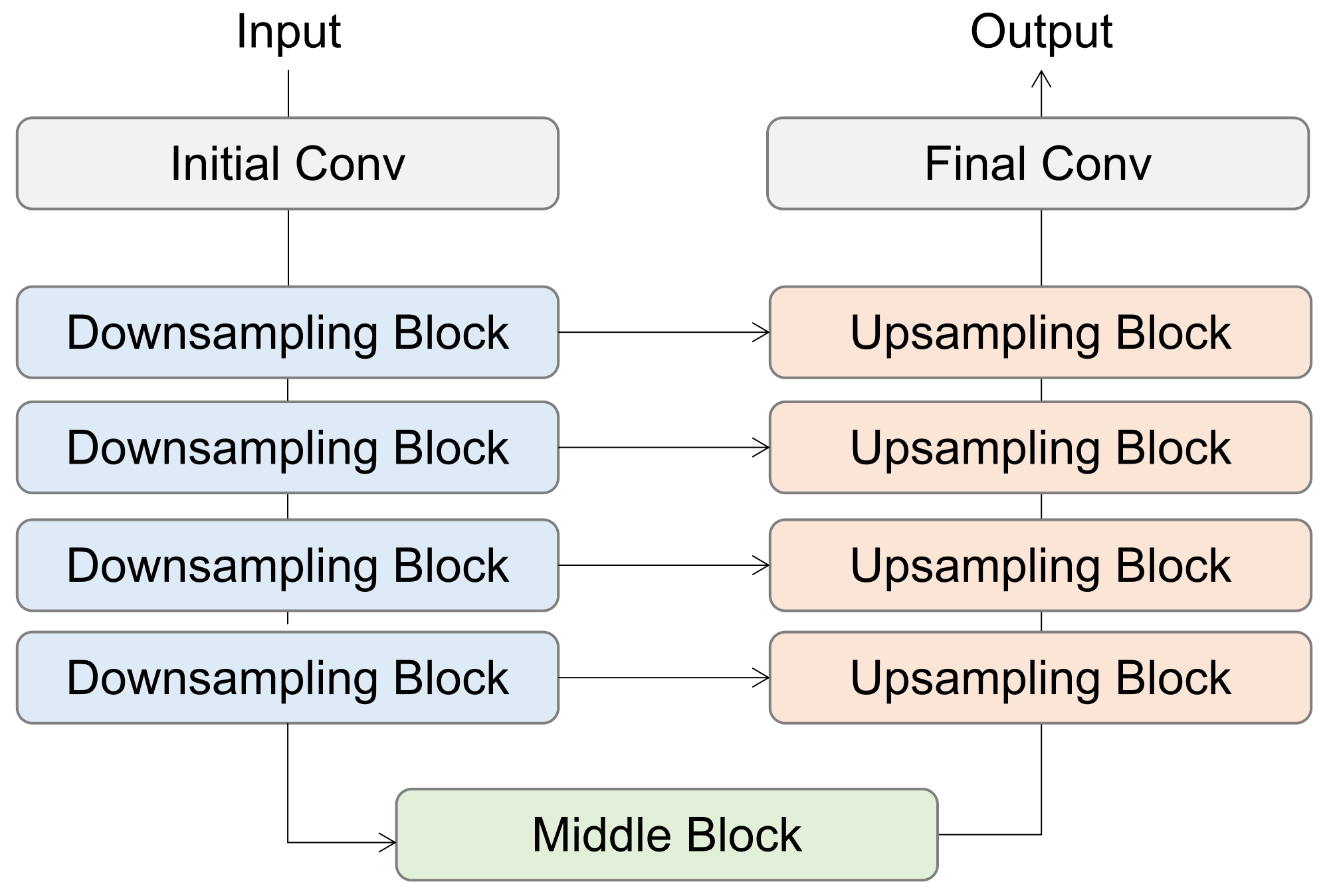}
        \caption{Outline of the architecture of NM-VQTSG.}
        \label{fig:}
    \end{subfigure}
    \\[0.7em]
    \begin{subfigure}[b]{0.23\textwidth} 
        \centering
        \includegraphics[width=\textwidth]{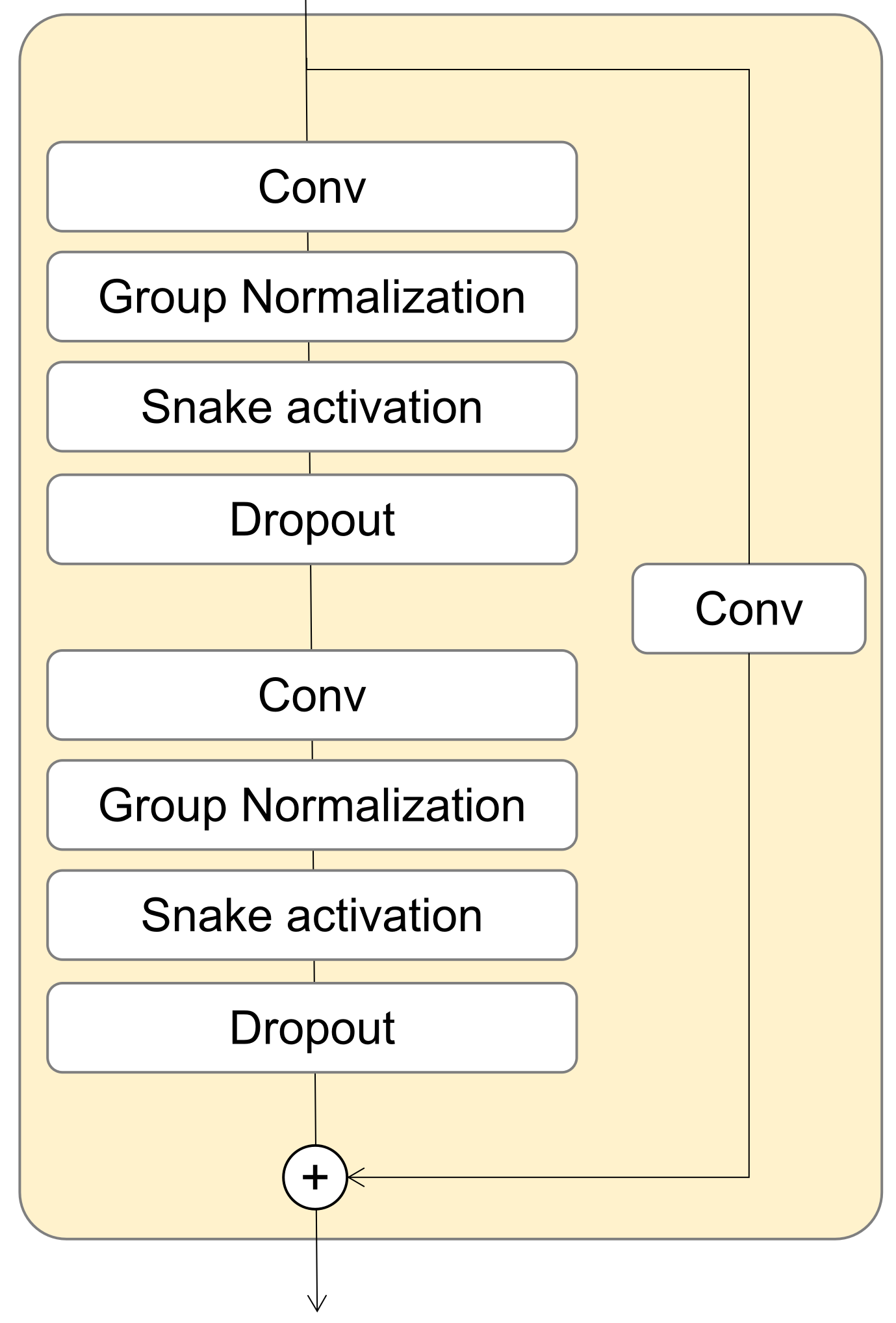}
        \caption{ResNet Block}
        \label{fig:}
    \end{subfigure}
    \hfill
    \begin{subfigure}[b]{0.2\textwidth} 
        \centering
        \includegraphics[width=\textwidth]{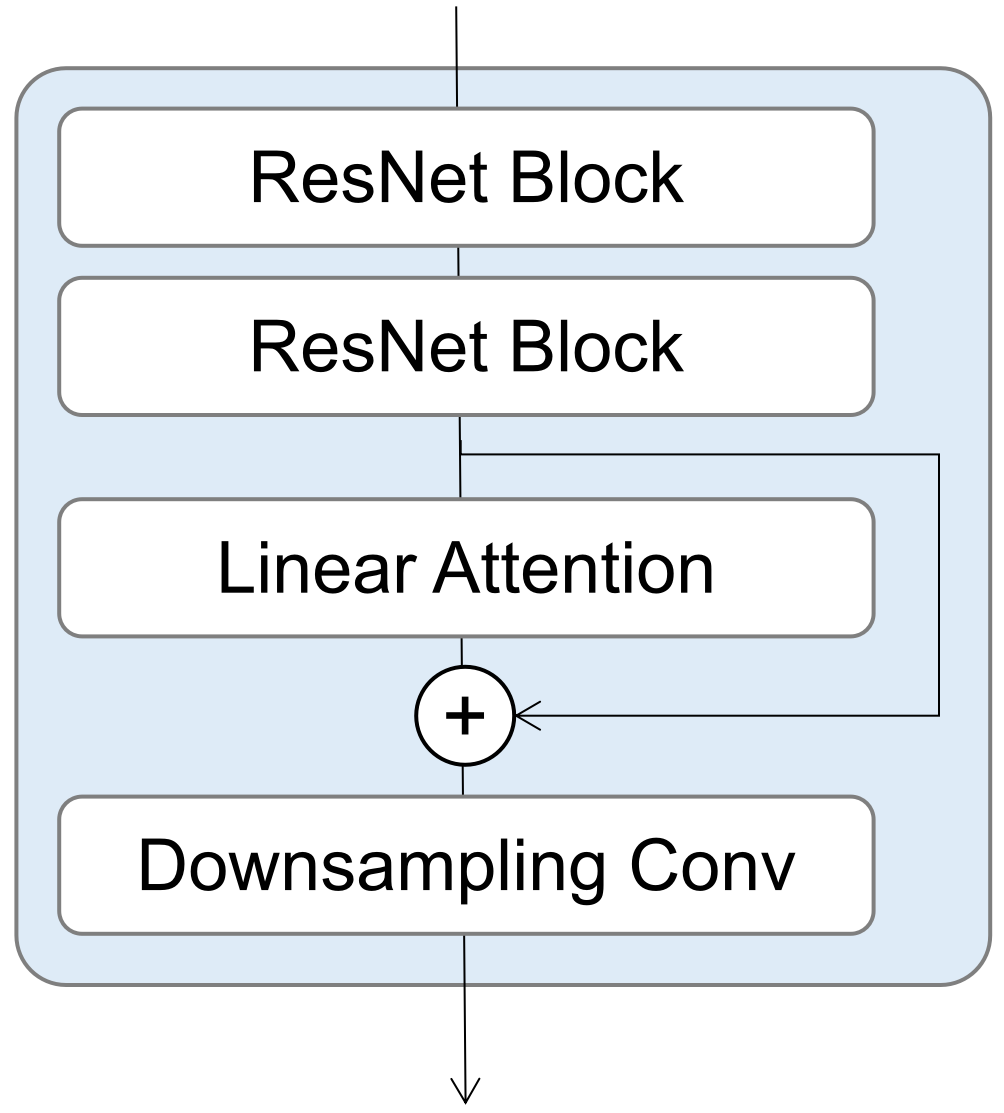}
        \caption{Downsampling Block}
        \label{fig:}
    \end{subfigure}
    \hfill
    \begin{subfigure}[b]{0.2\textwidth} 
        \centering
        \includegraphics[width=\textwidth]{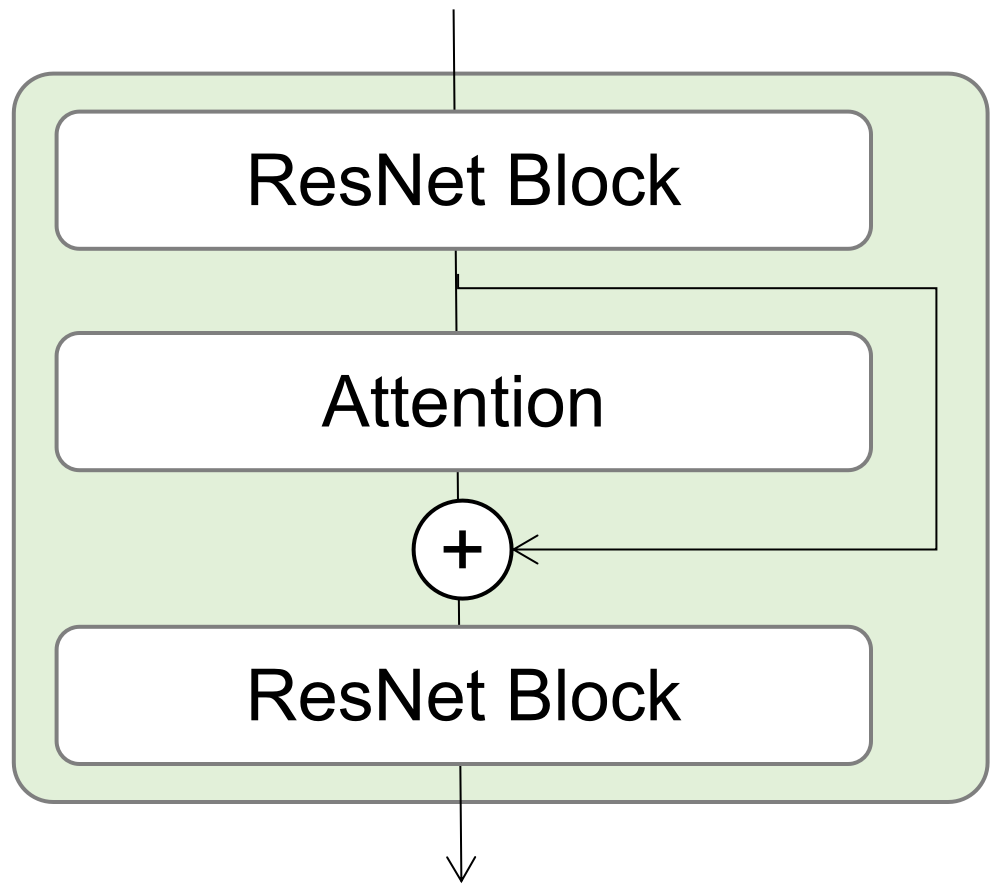}
        \caption{Middle Block}
        \label{fig:}
    \end{subfigure}
    \hfill
    \begin{subfigure}[b]{0.26\textwidth} 
        \centering
        \includegraphics[width=\textwidth]{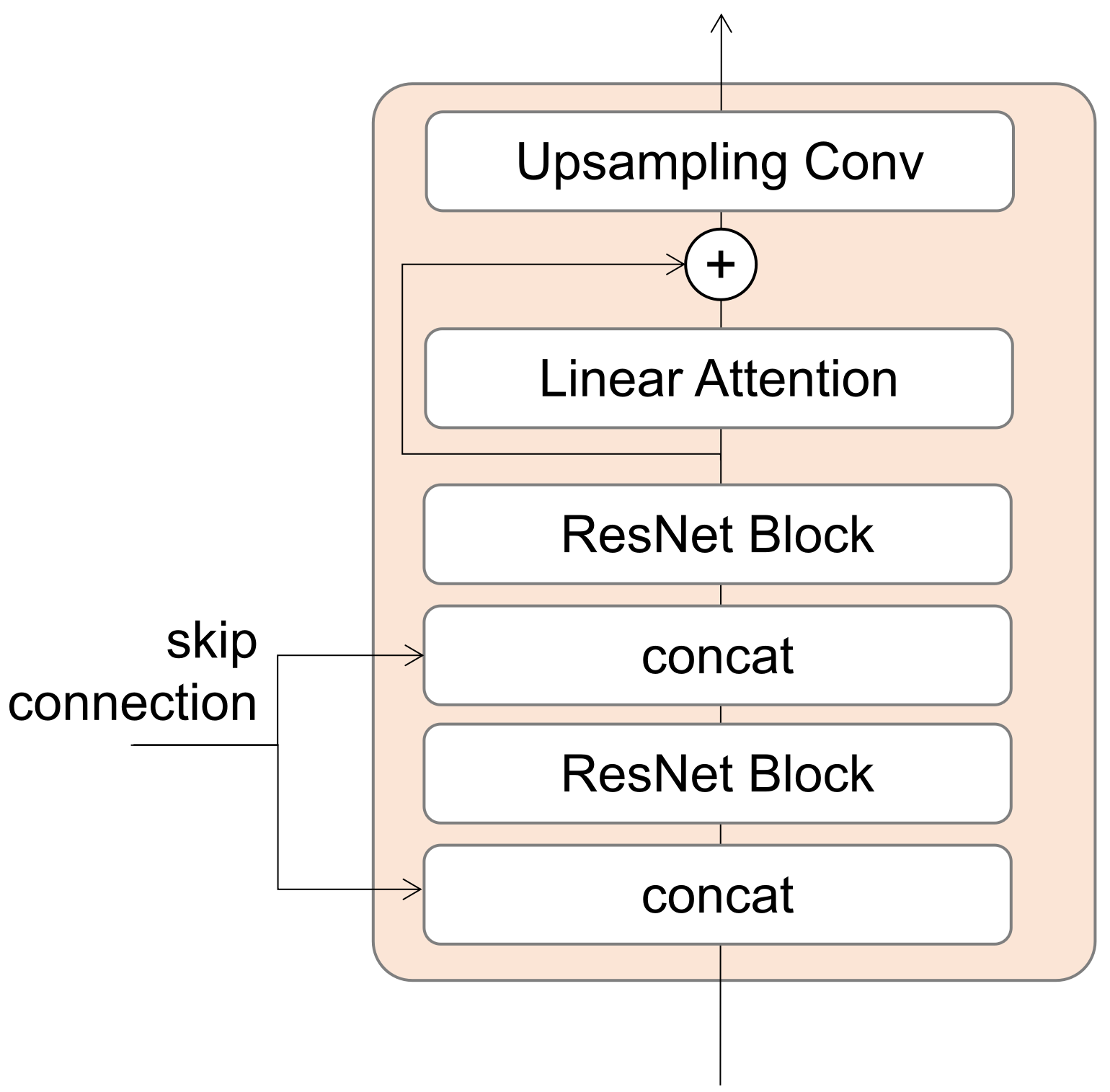}
        \caption{Upsampling Block}
        \label{fig:}
    \end{subfigure}
    \hfill
    \caption{Architecture of NM-VQTSG.
    Conv denotes a convolutional layer,
    Snake activation \cite{ziyin2020neural} has been shown highly effective at processing sinuous and periodic signals like audio \cite{kumar2024high}. We found that this activation function is effective for time series as well, especially for sinuous ones.
    At each downsampling convolutional layer (\texttt{Downsampling Conv}), the temporal length is halved while the embedding dimension is doubled. Similarly, at each upsampling layer, the process is reversed: the temporal length doubles, and the embedding dimension is halved.
    }
    \label{fig:architecture_NM}
\end{figure}

\end{document}